%% file: main.tex
\documentclass[10pt,twocolumn,letterpaper]{article}
 
\usepackage{cvpr}              % To produce the CAMERA-READY version
\usepackage{tabu}
\usepackage{booktabs}
\usepackage{tabularray}
\usepackage{multicol}
\usepackage{multirow}
\usepackage{caption}
\usepackage{graphicx}
\usepackage{listings}
\usepackage{makecell}
\usepackage{chapterbib}
\usepackage{enumitem}
\UseTblrLibrary{booktabs}

\usepackage{graphicx}
\usepackage{amsmath}
\usepackage{amssymb}

\usepackage{pifont}

\newcommand{\cmark}{\text{\ding{51}}}
\newcommand{\xmark}{\text{\ding{55}}}
\input{preamble}

\definecolor{cvprblue}{rgb}{0.21,0.49,0.74}
\usepackage[pagebackref,breaklinks,colorlinks,allcolors=cvprblue]{hyperref}

\def\paperID{10431} % *** Enter the Paper ID here
\def\confName{CVPR}
\def\confYear{2026}

\title{UniVerse: A Unified Modulation Framework for Segmentation-Free, Disentangled Multi-Concept Personalization}

\author{
Quynh Phung$^{1}$ \quad Sandesh Ghimire$^{2}$ \quad Minsi Hu$^{1}$ \\
Chung-Chi Tsai$^{2}$ \quad Jia-Bin Huang$^{1}$ \quad  \\
\\
$^1$~University of Maryland, College Park \quad
$^2$~Qualcomm Technologies, Inc.  \\
{\tt\small \{quynhpt,minsi,jbhuang\}@umd.edu} \quad
{\tt\small \{sghimire,chuntsai\}@qti.qualcomm.com}\\
\url{https://universe-personalization.github.io/}
}

\usepackage{pifont}

\begin{document}
% \maketitle

\twocolumn[{%
\renewcommand\twocolumn[1][]{#1}%
\maketitle
\vspace{-30pt}
\begin{center}
    \centering
    \captionsetup{type=figure}
    \includegraphics[width=1.\textwidth]{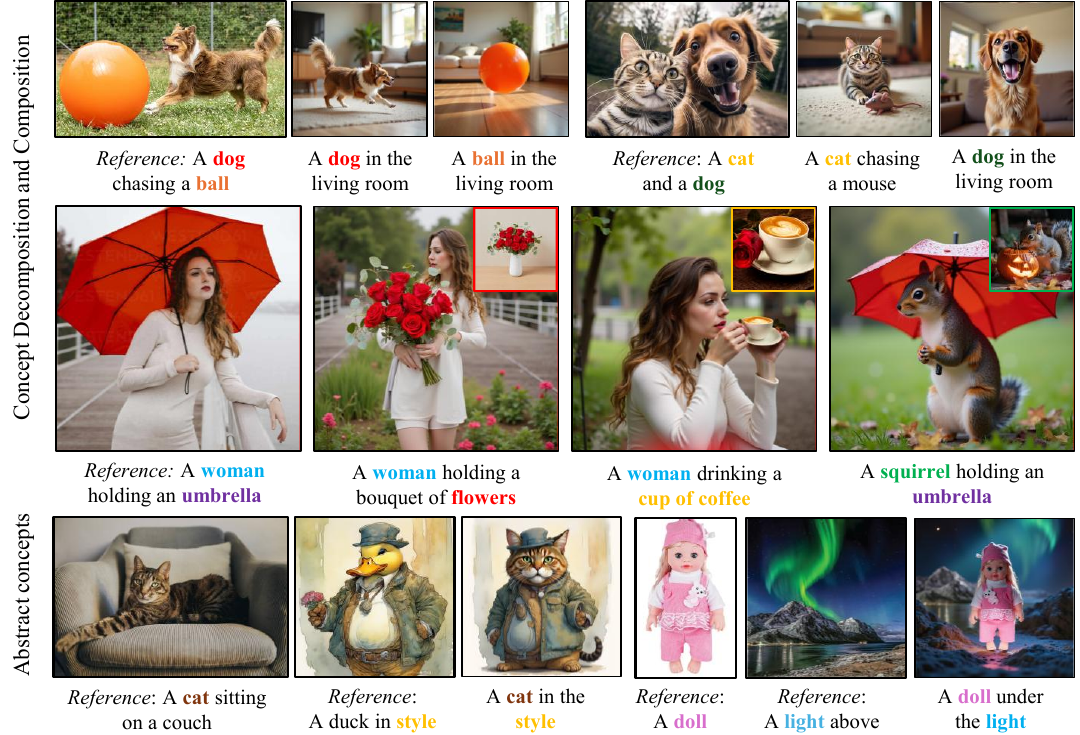}
    \caption{
    \textbf{Multi-concept customization with UniVerse}. 
Given a set of reference images and their corresponding text descriptions, our method seamlessly extracts relevant visual concepts and synthesizes new images by composing them, without requiring expensive model finetuning or segmentation.
Our approach effectively extracts concepts from objects with partial occlusion or abstract styles, and reliably preserves the distinct identities present in the reference images.
    }
    \label{fig:teaser}
\end{center}
}]

\input{sec/0_abstract}    
\input{sec/1_intro_new}
\input{sec/2_relatedwork}
\input{sec/3_method}

\input{sec/4_experiment}

\input{sec/5_conclusion}
{
    \small
    \bibliographystyle{ieeenat_fullname}
    \bibliography{main}
}


\end{document}

%% file: preamble.tex
\newcommand\mypara[1]{\vspace{1mm}\noindent\textbf{#1}.}

\newcommand{\Fref}[1]{Fig.~\ref{#1}}
\newcommand{\Tref}[1]{Table~\ref{#1}}

\newcommand{\R}[0]{\mathbb{R}}
\newcommand{\x}[0]{\times}

\definecolor{mypink}{rgb}{0.95, 0.95, 1.0}
\definecolor{myorange}{rgb}{0.98, 0.98, 0.98}
\definecolor{lost}{HTML}{ea4335}

\newcommand{\ndif}[1]{{\color{lost}-#1}}

\usepackage{comment}
\usepackage{tabularray}
\usepackage{tabu}
\usepackage{float}
\usepackage{placeins}

%% file: sec/0_abstract.tex
\begin{abstract}
Personalized visual understanding has advanced significantly, yet existing approaches struggle to localize and extract specific concepts when input images contain multiple objects. 
Many prior methods rely heavily on segmentation-based supervision or exhibit poor compositional generalization, limiting their ability to accurately disentangle and manipulate individual concepts. 
In this work, we propose UniVerse, a Unified Modulation Framework for segmentation-free, disentangled multi-concept personalization in diffusion transformers. 
Our method allows for composable and decomposable concept extraction, enabling fine-grained localization and representation of target objects without explicit segmentation masks. 
UniVerse learns to decompose complex scenes into concept-specific representations and then compose them in a unified manner, enabling robust personalization across diverse visual contexts. 
Through extensive experiments on multiple benchmarks, we demonstrate that UniVerse significantly outperforms state-of-the-art baselines in both localization accuracy and visual fidelity. 
Qualitative and quantitative results show that our approach can precisely extract target concepts in cluttered scenes, paving the way for more flexible, interpretable, and personalized visual generation and understanding.
\end{abstract}

%% file: sec/1_intro_new.tex
\vspace{-1em}
\section{Introduction}
\label{sec:intro}

The proliferation of text-to-image (T2I) generation models has unlocked remarkable capabilities in content creation. 
A key frontier in this field is personalization/customization: the ability to synthesize novel images featuring specific subjects, objects, or artistic styles provided by a user. 
This demand has spurred a rapid evolution of techniques. 
Early approaches, such as DreamBooth~\cite{ruiz2023dreambooth} and Textual Inversion~\cite{Textual_Inversion}, achieved high-fidelity personalization by \emph{fine-tuning} a model on a few reference images. 
While effective, this tuning-based paradigm is computationally expensive and requires a distinct optimization process for each new visual concept.

To address this, a second wave of \emph{tuning-free methods} emerged, including IP-Adapter~\cite{ye2023ip}, MIP-Adapter~\cite{MIP-Adapter}, PhotoMaker~\cite{li2024photomaker}, and PuLID~\cite{guo2024pulid}. 
These frameworks inject visual features from a reference image directly into the diffusion process, enabling flexible, zero-shot personalization without any per-subject training. 
However, many of these methods are built upon U-Net~\cite{U-Net} architectures with relatively weak text encoders.
This can limit their ability to handle complex compositions and nuanced semantic control, especially when compared to more recent, large-scale architectures.

The development of Diffusion Transformers~\cite{peebles2023scalable} (DiTs) marked a significant shift, offering superior scalability and a greater capacity for complex, multi-concept generation. 
This led to a new line of work, such as OmniGen~\cite{OmniGen}, UNO~\cite{UNO}, and DreamO~\cite{DreamO}, which leverage the transformer architecture to compose multiple distinct concepts within a single image. 
However, these unified transformer models often struggle with feature entanglement — where attributes from one subject leak into another — and with global feature injection, which degrades overall image quality. 

To overcome this entanglement problem, recent work has focused on modulation-based approaches, such as TokenVerse~\cite{garibi2025tokenverse}, Mod-Adapter~\cite{Mod-Adapter}, and XVerse~\cite{XVerse}. 
Instead of injecting broad visual features, these methods achieve finer control by modifying the text embedding signal itself. 
By calculating and applying modulation offsets to the text-conditioning stream, they can guide the generation process with high precision, preventing attribute leakage and enabling disentangled, high-quality personalization across multiple subjects.

However, these state-of-the-art modulation approaches introduce a critical new limitation. 
They predominantly require clean, pre-segmented reference images. 
This requirement severely curtails their practical utility, as real-world ``in-the-wild" use cases are dominated by complex, unsegmented photos. 
This in-the-wild challenge highlights a crucial, yet under-explored, aspect of personalization: the need for a precise reference prompt to guide concept extraction. 
Previous works~\cite{zhang2025survey} have largely overlooked this, either attempting to identify the desired concept using a generic single-word label (\eg, ``person") or bypassing the ambiguity entirely with segmentation masks, which are often unavailable.
Neither approach is sufficient. 
A simplistic prompt cannot disambiguate a specific subject from a cluttered background (\eg, ``the man on the left" in a group photo), and segmentation fails for many abstract concepts users wish to personalize, such as artistic styles, specific textures, or material properties. 
This dependency on manual pre-processing or overly simplistic prompts prevents these powerful models from generalizing to the very unconstrained scenarios where personalization is most valuable.

To bridge this crucial gap, we propose the Unified Modulation Framework---\textbf{UniVerse}---a novel framework for true segmentation-free, in-the-wild subject-driven generation. 
Unlike prior work that focuses on either visual-only conditioning or text-only modulation, or combines them loosely, our core innovation is a single, unified Reference Condition Extractor (RCE) that effectively extracts both visual conditional latents (for appearance) and textual conditional offsets (for semantics). 
Crucially, these two conditions are yielded by a single module and semantically aligned with the reference prompt, ensuring cohesive adherence to the image generation process. 
This dual-extraction pipeline allows our model to automatically decompose complex reference images, identify, and disentangle multiple concepts—from distinct object identities to abstract artistic styles. 
These disentangled concepts can then be flexibly composed to synthesize new, complex scenes.
\Fref{fig:teaser} illustrates the decomposition ability of our approach on in-the-wild reference images while maintaining the photorealism of the generated images in new contexts or when combining multiple concepts.

In this work, our proposed solutions include: 
\begin{enumerate*}[(i)]
\item the Reference Condition Extractor, the first framework to extract semantically-aligned visual and textual conditions from a single module, guided by a reference prompt, to enable robust multi-concept decomposition and composition.
\item a two-stage training strategy: the extractor is first supervised on a reference segmentation dataset, teaching it to accurately localize concepts from a prompt, and is then jointly trained for the full generation task, resulting in significantly improved generalization;
    %a two-stage training strategy. The extractor is first supervised on a reference segmentation dataset, which teaches the model to accurately localize concepts from a prompt before being jointly trained for the full generation task, significantly improving generalization.    
\item UniVerseBench, a new benchmark of multi-concept reference images designed to rigorously evaluate prompt-guided concept decomposition, testing a model's ability to disambiguate and extract the correct concept—something existing benchmarks do not adequately cover.
    
    % to rigorously evaluate the nuanced task of prompt-guided concept decomposition, we introduce UniVerseBench, a new benchmark featuring multi-concept reference images. This benchmark specifically tests a model's ability to disambiguate and extract the correct concept based on a textual prompt, a scenario that existing benchmarks do not adequately cover.
\end{enumerate*}

%% file: sec/2_relatedwork.tex
\section{Related Work}
\label{sec:relatedwork}
% token verse: no code - https://token-verse.github.io/
% xverse: https://github.com/bytedance/XVerse
% UNO: https://github.com/bytedance/UNO
% dreamo: https://github.com/bytedance/DreamO
% Omnigen2: https://github.com/VectorSpaceLab/OmniGen2
% Omnigen: https://github.com/VectorSpaceLab/OmniGen
% MS-Diffusion:  https://github.com/MS-Diffusion/MS-Diffusion
% MIP-adapter: https://github.com/hqhQAQ/MIP-Adapter

\mypara{Subject-driven generation}
The main challenge in subject-driven generation is maintaining a subject’s visual identity while allowing flexible text-based editing. 
Early works achieve this through fine-tuning approaches, where pre-trained diffusion models are adapted to new subjects using a few reference images (\eg, DreamBooth~\cite{ruiz2023dreambooth}, Textual Inversion~\cite{Textual_Inversion}). 
Later methods introduce more efficient tuning-free strategies that inject subject features directly into the diffusion process without model retraining. 
Representative works include IP-Adapter~\cite{ye2023ip}, MIP-Adapter~\cite{MIP-Adapter}, SSR-Encoder~\cite{zhang2024ssr}, PhotoMaker~\cite{li2024photomaker}, and PulID~\cite{guo2024pulid}, which employ pre-trained image encoders and cross-attention mechanisms to transfer visual representations, supporting zero-shot or low-resource personalization across subjects.

\mypara{Multi-concept control and feature injection}
Scaling personalization to multiple concepts or subjects introduces challenges of feature disentanglement and spatial control. 
A large body of work leverages attention-based conditioning to manage multiple subjects and modalities within a unified framework, such as OmniGen~\cite{OmniGen}, OmniGen2~\cite{OmniGen2}, DreamO~\cite{DreamO}, UNO~\cite{UNO}, and MS-Diffusion~\cite{MS-Diffusion}. 
Other methods focus on localized editing and grounding, providing fine-grained spatial control using auxiliary cues such as segmentation maps, bounding boxes, or depth information. 
Examples include Break-a-scene~\cite{avrahami2023break} and SeedEdit~\cite{shi2024seededit}, which enable natural, in-the-wild customization with strong spatial grounding.

\mypara{DiT-Based Modulation Approaches}
The recent shift from UNet-based architectures to Diffusion Transformers (DiTs)~\cite{peebles2023scalable} has enabled more structured and scalable conditioning mechanisms. 
In these models, Adaptive Layer Normalization (AdaLN)~\cite{AdaLN} provides a clean way to modulate the generation process via learned scale and bias terms. 
Building on this principle, frameworks such as TokenVerse~\cite{garibi2025tokenverse} and XVerse~\cite{XVerse} demonstrate that modulating the text-token space can effectively control multiple concepts without explicit masks or segmentation. 
These token-based modulation techniques offer fine-grained, disentangled personalization, paving the way for segmentation-free and highly controllable subject customization, which motivates the design of the proposed framework.

\Tref{tab:related_works} summarizes the ability of our method, UniVerse, with existing personalized image generation models.
\begin{table}[t!]
% \vspace{-0.2cm}
\caption{UniVerse supports both concept decomposition and subsequent composition, as well as handling both multiple and abstract concepts.}
\label{tab:related_works}
\vspace{-0.5em}
\centering
\footnotesize
\begin{tblr}{width=\linewidth,colspec={@{}X[1.7,l]X[1.1,c]X[1.2,c]X[1.2,c]X[1.2,c]},row{8}={mypink},stretch=0}
\toprule
\SetCell[r=2]{l}Model & Concept & Concept & Multiple & Abstract \\ 
& Comp. & Decomp. & Concepts & Concepts \\
\midrule
XVerse~\cite{XVerse} & $\cmark$ & $\xmark$ & $\cmark$ & $\cmark$ \\ 
UNO~\cite{UNO} & $\cmark$ & $\xmark$ & $\cmark$ & $\cmark$\\
DreamO~\cite{DreamO} &  $\cmark$ & $\xmark$ & $\cmark$ & $\cmark$\\ 
OmniGen~\cite{OmniGen} &$\cmark$  & $\xmark$ & $\cmark$  & $\xmark$ \\ 
\mbox{MS-Diffusion~\cite{MS-Diffusion}} &$\cmark$  & $\xmark$ & $\xmark$ &$\cmark$\\
\mbox{\textbf{UniVerse (Ours)}} &  $\cmark$ & $\cmark$ & $\cmark$ & $\cmark$ \\ 
\bottomrule
\end{tblr}
\vspace{-0.7em}
\end{table}

%% file: sec/3_method.tex
\section{Methodology}

In this section, we first review the DiT modulation used for personalization/customization in previous works. 
We then discuss the UniVerse framework, which introduces a novel module to extract both visual and textual additional conditions.
The following section details our approach to dataset preparation for handling in-the-wild reference images.

% \footnote{Other spatial representations, segmentation maps, edges, vector graphics, and ASCII art, can also be extracted. We focus on using bounding boxes as the spatial layout in this work.}
% , segmentation maps, edges, vector graphics, and ASCII art. 
% Given the generated layout, we propose novel attention-refocusing losses to guide the pre-trained text-to-image models. 
% We first present the background in \Sref{sec:method_prelim}. 
% We then derive our proposed loss aiming to improve the alignment with respect to the layouts in \Sref{sec:method_txt2img} 
% \Sref{sec:method_txt2img} derives our proposed loss aiming to improve the alignment with respect to the layouts. 
% In \Sref{sec:method_layout}, we describe how we prompt LLMs to generate layouts. 

\begin{figure*}[t]
    \vspace{-1em}
    \centering
    \includegraphics[width=0.57\linewidth]{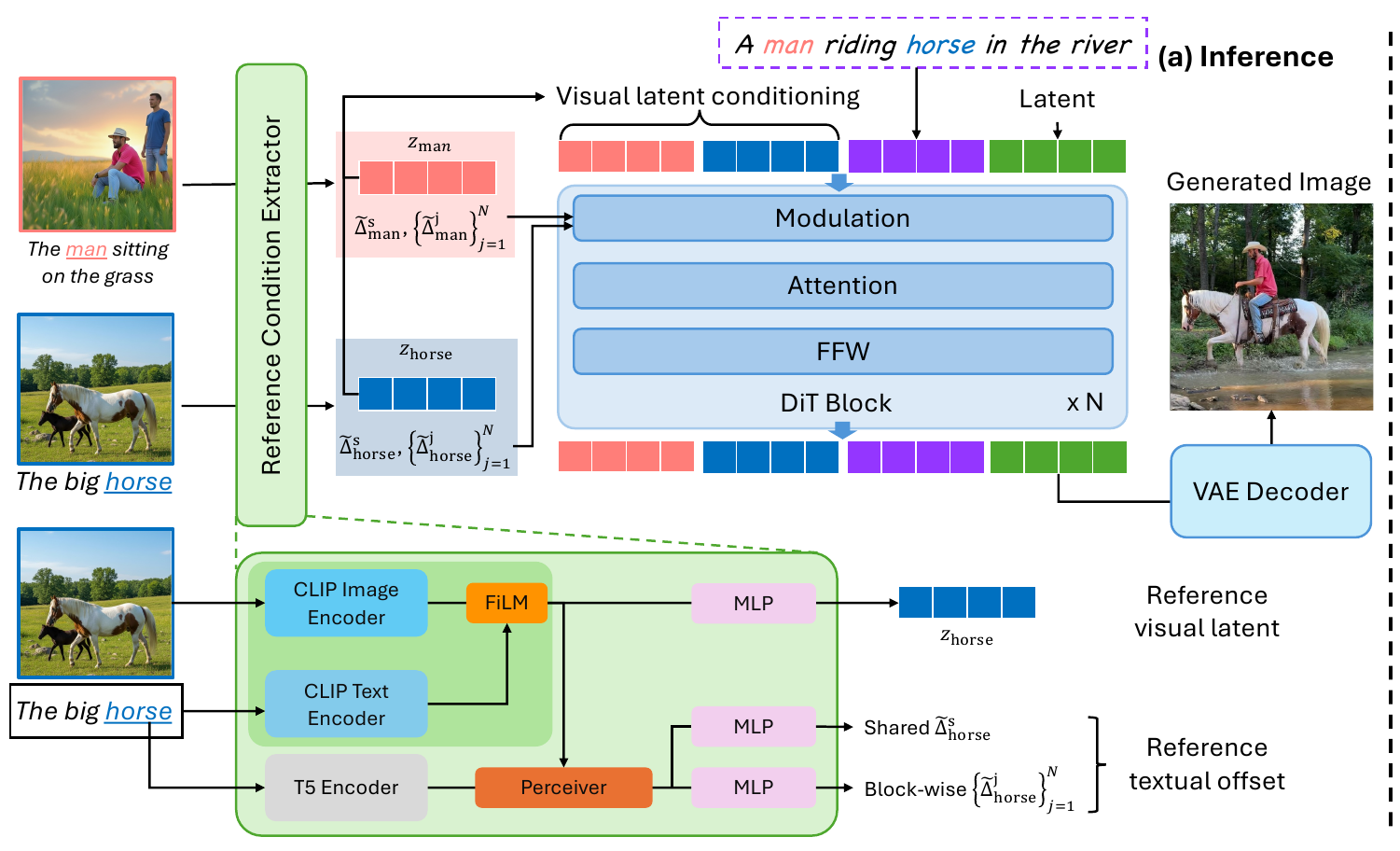}
    \includegraphics[width=0.42\linewidth]{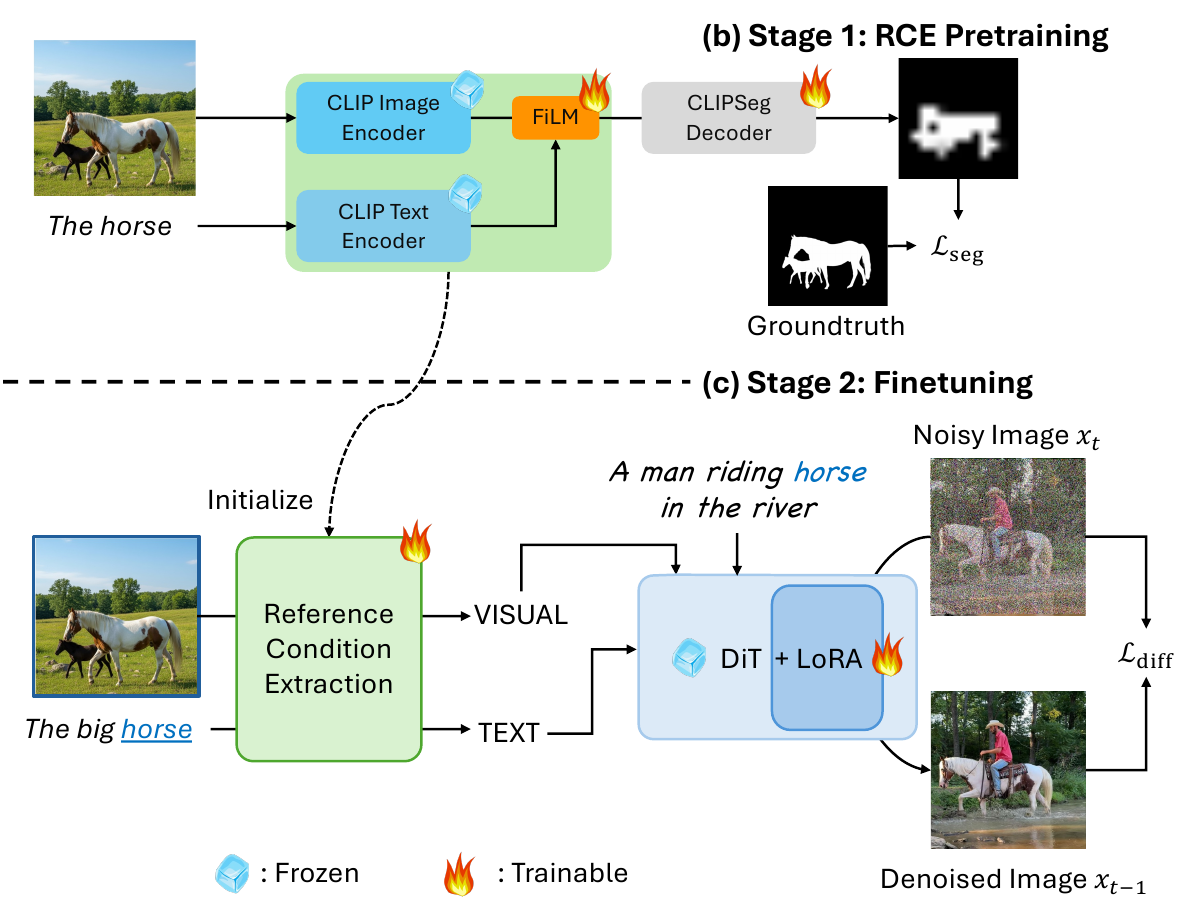}
    \vspace{-0.5em}
    \caption{
    \textbf{Our proposed UniVerse Framework to generate personalized images from in-the-wild reference images.} \textit{(a) Inference:} 
    The Reference Condition Extractor (RCE) extracts both visual and textual references. The two features are extracted from CLIP~\cite{CLIP} and T5~\cite{T5} encoders with additional modules to adapt to DiT blocks. The textual reference includes a shared vector $\tilde{\Delta}^s$ modulates all DiT blocks and block-wise vector sets $\{\tilde{\Delta}^j\}_{j=1}^N$. The visual condition $z_\text{ref}$ is used as an additional latent to deeply control the generated images. 
    \textit{(b) Stage 1 - RCE Pretraining:} 
    The segmentation head was added to facilitate training on a large-scale dataset. Binary cross-entropy (BCE) is used as the segmentation loss. 
    \textit{(c) Stage 2 - Finetuning:} the FiLM~\cite{perez2018film} is continue finetuned with other blocks on multi-concept dataset. Here, LoRA~\cite{LoRA} is added to the DiT and the whole process is trained with diffusion loss $\mathcal{L}_\text{diff}$.}
    \label{fig:overall}
    \vspace{-0.7em}
\end{figure*}

\subsection{Preliminaries}
\label{sec:method_prelim}

Diffusion Transformers (DiTs) have become the foundational architecture for scalable image synthesis, replacing UNets~\cite{saharia2022photorealistic, ramesh2022hierarchical, rombach2022high, SDXL} in models such as Stable Diffusion 3~\cite{Stable-Diffusion-3}. DiTs employ a unique, high-level mechanism for integrating conditioning information (\eg, the CLIP~\cite{CLIP} text prompt embedding $f_T(p)$ and the timestep $t$), known as modulation. This is achieved through Adaptive Layer Normalization (AdaLN)~\cite{AdaLN}, where a Multi-Layer Perceptron (MLP) processes the inputs to generate a conditioning vector $y$:
\begin{align}
    y = \text{MLP}(t, f(p)). 
\end{align}

This vector is then split into scale ($\alpha$) and shift ($\beta$) terms that dynamically modulate the network's activations, effectively integrating semantic control separately from the primary data flow. 
% This clean separation of semantic control minimizes artifacts and enhances editability. Leveraging this high-level semantic space, subsequent research focused on personalization. 
TokenVerse~\cite{garibi2025tokenverse} pioneered injecting personalized identity features directly into this modulation pathway by learning a personalized vector offset ($\Delta$) per text token rather than using the same vector $y$ to modulate all tokens. Building on this, XVerse~\cite{XVerse} achieved tuning-free (zero-shot) subject-specific control by using a universal adapter to generate an offset $\tilde{\Delta}^i$ for $i-$th token from its corresponding reference image $I_i$. Here, the new offset is added to the modulation vector as: $\tilde{y}_i=y+\tilde{\Delta}_i$

% We propose UniVerse framework which focuses on generating a better offset $\Delta_{cross}$ from reference images  segmentation-free \textbf{Segmentation-Free Disentangled Modulation (SFDM)} framework extends this powerful modulation lineage to handle localized, multiple concepts within a single, unsegmented image in a zero-shot manner.

\subsection{UniVerse Framework}
We improve the tuning-free approach in generating $\tilde{y}$ for the modulation in DiTs. Besides the reference image $I_i$ and corresponding token $p_i$, our model accepts the reference prompt $r_i$ describing the reference object in the context of the reference image. It will help the model know which concept to extract from the reference image. In some cases, the reference prompt contradicts the prompt token in the full image. For example, in ~\Fref{fig:overall}, where ``sitting on the grass" is the action of the man in the reference image, it moves to ``riding a horse" in the final prompt. 

Our main pipeline, illustrated in~\Fref{fig:overall}, includes the Reference Condition Extractor (RCE), which generates both textual and additional visual conditions for DiTs during image generation. 
While the textual condition is in the form of modulated offset $\tilde{\Delta}_i$, the visual latent $z_\text{ref}$ is appended to all latent inputs as additional conditions for denoising the image. 
The procedure for obtaining the two conditions is described in the following paragraphs.

\mypara{Visual Reference Latents} 
We leverage both CLIP~\cite{CLIP} image and text encoders to extract reference-image and prompt features, respectively. 
The visual features are then modulated by the textual features via Feature-wise Linear Modulation (FiLM)~\cite{perez2018film}, where it modifies each visual vector to remove unnecessary information. 
Given the visual features $\mathbf{F} = f_V(I) \in \R^{N \x D}$ and textual features $x=f_T(r) \in \R^D$ where $N$ is the number of visual tokens and $D$ is the feature dimension, the modulated visual features are modifying $f_V(I)$ as:
\begin{align}
    \text{FiLM}(\mathbf{F}_j, x) = g_i(x) \mathbf{F}_j + h_i(x)
\end{align}
This function modulates each vector $\mathbf{F}_j$ by shift and scale derived from functions $g(.)$ and $h(.)$. The following MLP layer projects the conditions to the DiT latent space as $z_\text{ref}$.

% NOTE WE PUT FIGURE FROM EXPERIMENT HERE TO DISPERSE FIGURERS
% FIGURE APPEARS ON NEXT PAGE
% MOVE THIS FIGURE AROUND AS FIT
% single subject comparisons
\begin{figure*}[t]
    \centering
    % \vspace{-1em}
    \includegraphics[width=0.9\linewidth]{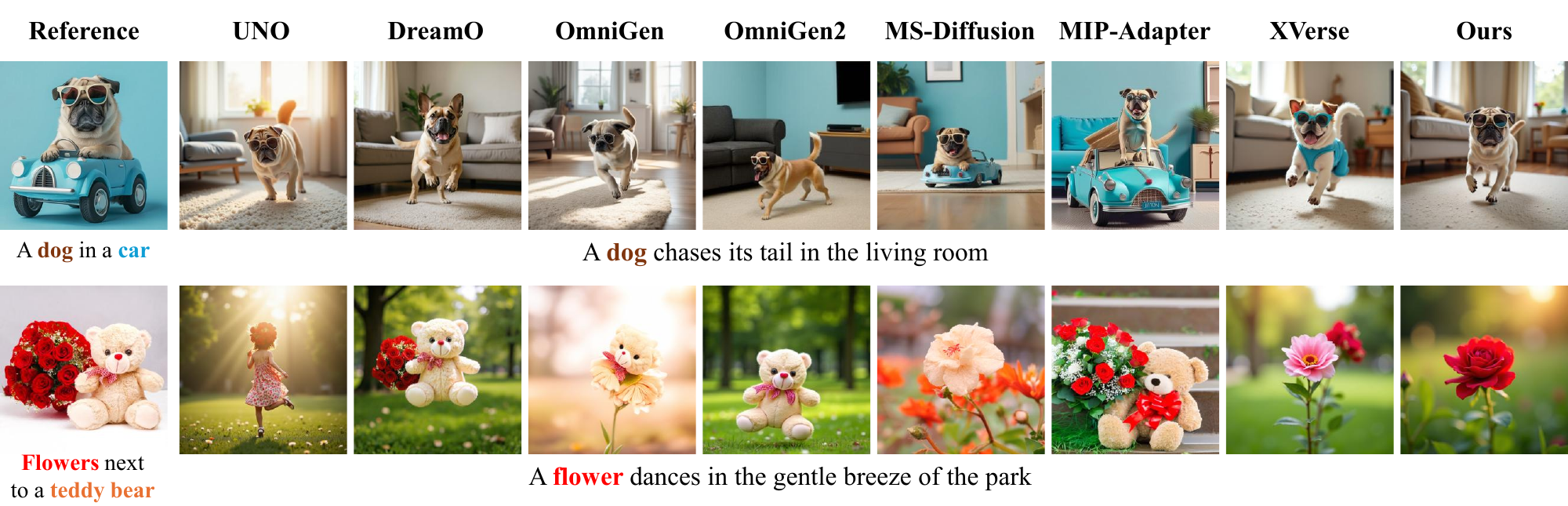}
    % \vspace{-.1cm}
\caption{\textbf{Concept extraction comparisons for single-subject generation}. Each row depicts a reference image (left) and images containing a concept from the reference image, generated by UNO~\cite{UNO}, DreamO~\cite{DreamO}, OmniGen~\cite{OmniGen}, OmniGen2~\cite{OmniGen2}, MS-Diffusion~\cite{MS-Diffusion}, MIP-Adapter~\cite{MIP-Adapter}, XVerse~\cite{XVerse}, and our method (UniVerse). For the first row, MS-Diffusion, MIP-Adapter, and XVerse suffer from concept leakage, DreamO fails to preserve the sunglasses on the dog, and UNO, OmniGen, and OmniGen2 have issues with subject fidelity. Meanwhile, our method seamlessly extracts the concept of "dog" while preserving subject fidelity and details. 
For the second row, none of the baselines completely disentangle the concepts of "flowers" and "teddy bear," except for our method and XVerse; only our method preserves the specific flower from the concept image.}
    \label{fig:qual_single}
    % \vspace{-0.7em}
\end{figure*}

% FIGURE HERE TO DISPERSE
% multi subject comparisons
\begin{figure*}[t]
    % \vspace{-1em}
    \centering
    \includegraphics[width=0.9\linewidth]{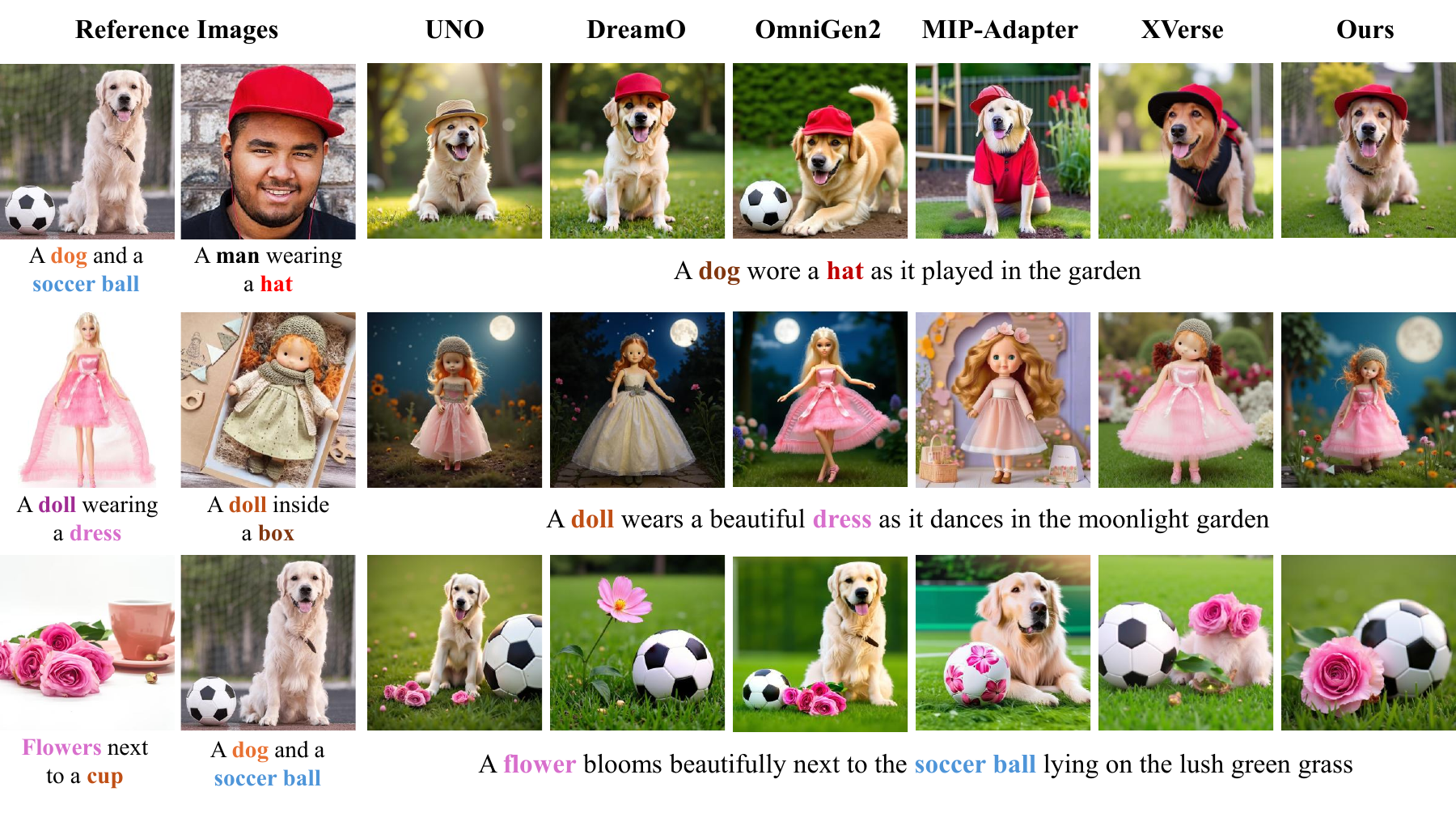}
    % \vspace{-.3cm}
    \caption{
    \textbf{Concept extraction and composition comparisons for multi-subject generation}. 
    Each row depicts reference images (left) and images containing a concept from the reference images, generated by UNO~\cite{UNO}, DreamO~\cite{DreamO}, OmniGen2~\cite{OmniGen2}, MIP-Adapter~\cite{MIP-Adapter}, XVerse~\cite{XVerse}, and our method (UniVerse). For the first row, XVerse, OmniGen2, and MIP-Adapter suffer from leakage while UNO composes the wrong hat. 
    Between DreamO and our method, our method best preserves the characteristics of the reference concepts, even retaining the dog's black collar. The second row illustrates a more difficult task where both references contain a doll---most methods suffer from poor concept extraction, preservation, or composition. 
    Meanwhile, our method can disentangle concepts and compose them effectively. For the third row, other methods continue to suffer from poor extraction or composition. 
    DreamO can extract relevant concepts but generates the wrong flower, whereas our method is the only one to correctly complete the task.}
    \label{fig:qual_multi}
    % \vspace{-0.7em}
\end{figure*}

\mypara{Textual Reference Offset} Following XVerse~\cite{XVerse}, we inject visual features into the T5~\cite{T5} embeddings of prompt token $p_i$ via a Perceiver~\cite{jaegle2021perceiver} layer. 
However, instead of using the CLIP image features directly, we leverage modulated visual features with non-essential information removed. At the end, we learn two modulation offsets for each reference token $p_i$, one shared $\tilde{\Delta}^\text{s}_i$ between blocks and a specific $\tilde{\Delta}^j_i$ for each block. The final modulation vector for $i-$th token at block $j$ is $\tilde{y}^j_i = y + \tilde{\Delta}^\text{s}_i + \tilde{\Delta}^j_i$

\subsection{Two-stage Training Pipeline}
Our proposed training approach is shown in \Fref{fig:overall} (b) and (c). 
There are two stages: We first pretrain the FiLM~\cite{perez2018film} module on a large-scale dataset and then finetune with other modules in the second stage on our multi-concept dataset.

In the first stage, we train the single FiLM layer alone with output from the CLIP encoders. 
The process is supervised by reference instance segmentation, where an additional CLIPSeg Decoder~\cite{CLIP-Seg} is added to predict a coarse segmentation mask conditioned on the text. 
We use binary cross-entropy as our loss $\mathcal{L}_\text{seg}$. In this stage, only the FiLM and the CLIPSeg Decoder are trained. 

In the second stage, we train the entire pipeline, including the Reference Condition Extractor (RCE) and DiT, on the reference-image-generation task. 
All encoder networks remain frozen, while the remaining components of the RCE are trainable.
%Except for all encoders are kept frozen, other modules are trainable in RCE. 
The FiLM module is initialized from the previous stage, while other layers are trained from scratch. 
We add low-rank parameters (LoRA~\cite{LoRA}) to DiT and train them, while modulating their normalization parameters with learned offsets from RCE. The standard diffusion loss $\mathcal{L}_\text{diff}$ is used on the noise prediction. 
% For efficient training, only one reference image is used in each training sample.

% \subsection{Data creation pipeline}
% \begin{comment} % REMOVE THIS WHEN FIGURE COMPLETE
% \begin{figure}[t]
% \vspace{-0.5em}
%     \centering
%     \includegraphics[width=\linewidth]{figures/.pdf}
%     \vspace{-0.5cm}
%     \caption{\tb{data pipeline.} .}
%     \label{fig:data}
%     \vspace{-0.5cm}
% \end{figure}
% \end{comment}

%% file: sec/4_experiment.tex
\section{Experiment}
\label{sec:experiment}

% just our model's images
\begin{figure*}[t]
    \centering
    \includegraphics[width=\linewidth]{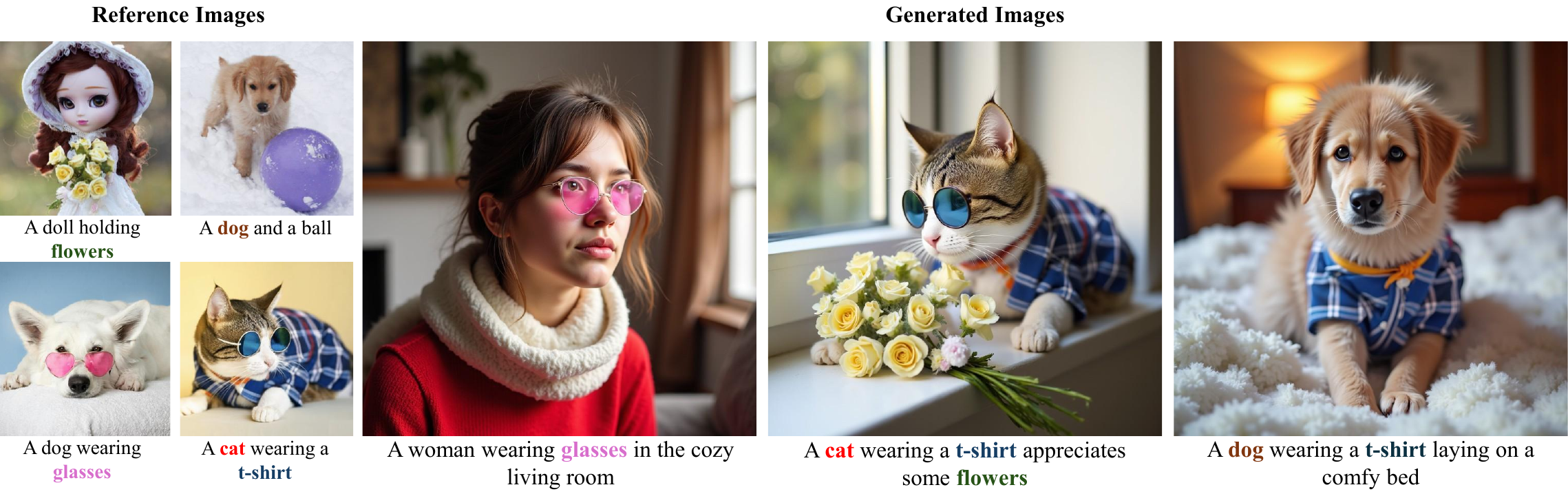}
    \vspace{-2em}
    \caption{\textbf{Qualitative results}. On the left of each row, we have four reference images, each consisting of multiple concepts. 
    On the right, we show three generated images produced by our method, demonstrating its ability to seamlessly extract and combine concepts from multiple reference images without explicit segmentation. Refer to the supplementary materials for additional results.}
    \label{fig:qual_ours}
    \vspace{-0.7em}
\end{figure*}

\begin{figure}[t]
    \centering
     \vspace{-1.5em}
    \includegraphics[width=\linewidth]{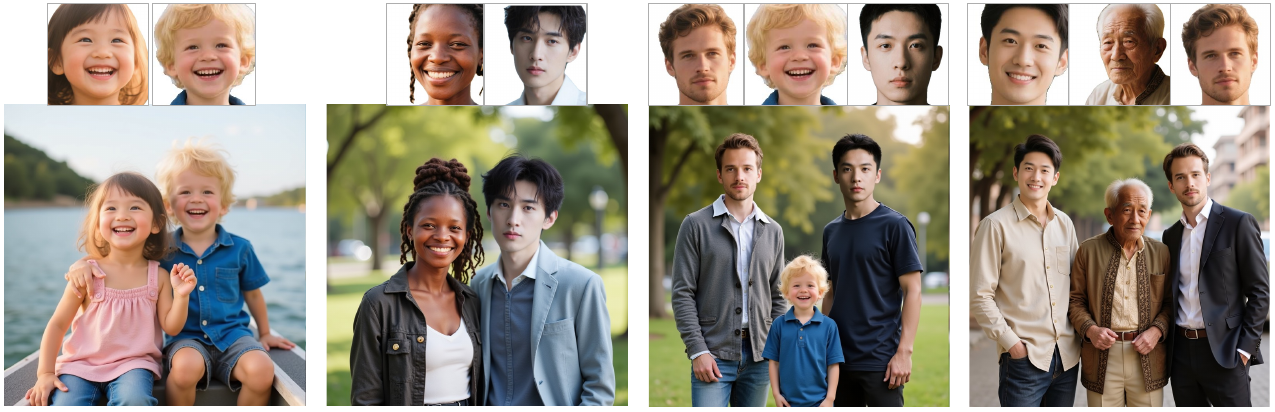}
    \vspace{-1em}
    \caption{\textbf{Multi-person composition}. UniVerse successfully preserves individual identities during multi-human generation.}
    \label{fig:human}
    \vspace{-1em}
\end{figure}

\begin{figure}[t]
    \centering
     \vspace{-0.5em}
    \includegraphics[width=\linewidth]{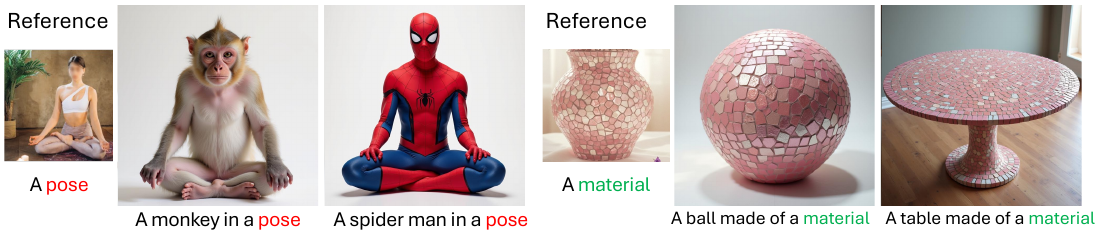}

    \vspace{-1em}
     \caption{\textbf{Generalization to abstract concepts}.UniVerse successfully disentangles and composes abstract attributes such as pose and material.}
     \label{fig:pose}
    \vspace{-2em}
\end{figure}

\begin{figure*}[t]
    \centering
     \vspace{-3em}
    \includegraphics[width=1.0\textwidth]{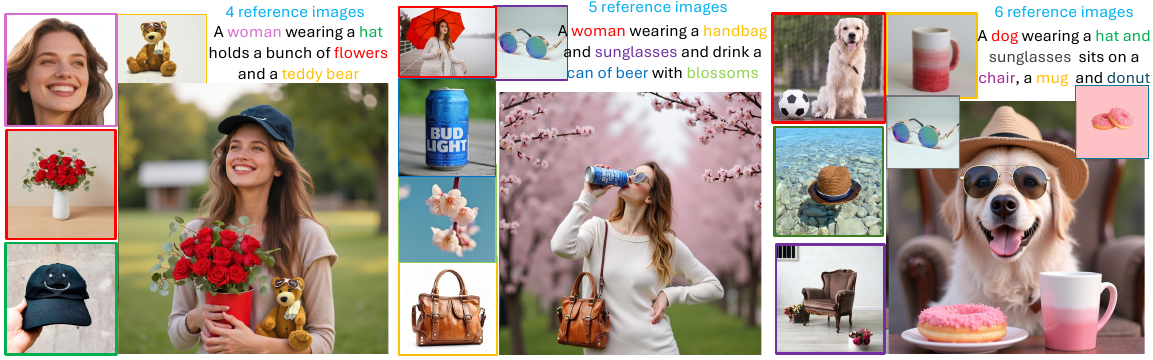}
     \caption{\textbf{Multiple objects.}.UniVerse effectively disentangles and composes up to six distinct objects while maintaining high identity fidelity for each subject.}

    \label{fig:multi_obj}
    \vspace{-1em}
\end{figure*}

\begin{figure}[H]
    \centering
     \vspace{-2em}
    \includegraphics[width=\linewidth]{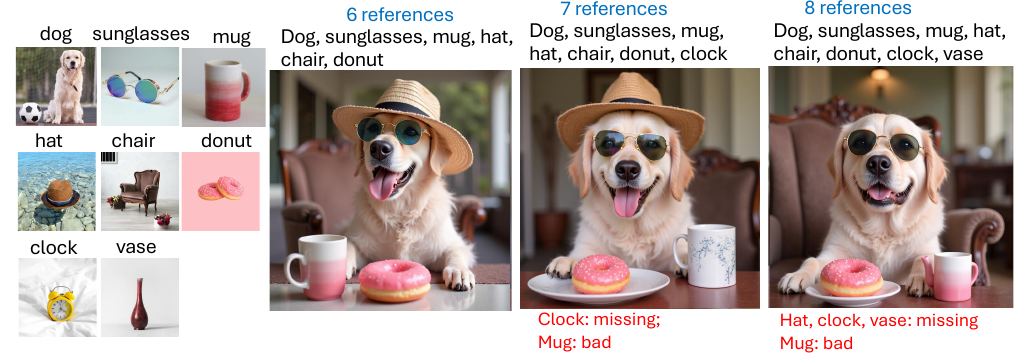}
    \vspace{-1em}
     \caption{\textbf{Compositional capacity}.UniVerse maintains identity fidelity for up to 6 subjects; however, exceeding this threshold can result in identity crosstalk or missing instances.}
  
    \label{fig:num_ref}
    \vspace{-2em}
\end{figure}

We conduct both quantitative and qualitative evaluations, demonstrating that UniVerse surpasses existing methods in accurately extracting multiple visual concepts from reference images and effectively integrating them to generate new, coherent images.
\textbf{We will open-source our code and pretrained model for reproducible research.}

\subsection{Implementation Details} 

\mypara{Training Datasets} 
We train our model using publicly available datasets and our own curated datasets. 
For the first stage, we pretrain our Reference Condition Extractor (RCE) on PhraseCut~\cite{PhraseCut}, a large-scale dataset for reference image segmentation. 
For the second stage, inspired by prior work~\cite{UNO, XVerse}, we build our own conceptual dataset using images from UNO-1M~\cite{UNO}. 
With a limited number of multi-concept samples, inspired by prior work~\cite{XVerse}, we also horizontally combine the reference images to help the model learn to extract the correct concepts. In our setting, we call this augmentation technique Cross-Reference. 
Details of the dataset creation pipeline are available in the supplementary material. 

\mypara{Benchmarks} We evaluate our model and baselines on two public benchmarks: DreamBench++~\cite{DreamBench++} and XVerseBench~\cite{XVerse}. While DreamBench++ is designed to evaluate single-concept personalization in text-to-image generation, XVerseBench extends the evaluation to multi-concept composition and fine-grained attribute control.

To further evaluate models' ability to disentangle co-occurring visual concepts within the same reference images, we propose a new benchmark, UniVerseBench. 
The dataset for UniVerseBench consists of 20 reference images and 200 distinct prompts to evaluate single- and multi-subject image generation. 
Unlike previous benchmarks, UniVerseBench focuses on \emph{object decomposition} from reference images. 
Each reference image consists of two co-occurring subjects, challenging models to extract the correct concept under these conditions. 

% figures for quantitative results:
% XVerseBench
\begin{table*}[t]
\caption{Quantitative comparison of single-subject and multi-subject personalization performance on XVerseBench. \textbf{Bold} represents best performance, \underline{underline} represents second best.}
\vspace{-0.3cm}
\label{tab:XVerseBench}
\centering
\footnotesize
\begin{tblr}{width=\linewidth,colspec={@{}X[2.2,l]|*{5}{X[0.8,c]}|*{5}{X[0.8,c]}|X[1.1,c]},stretch=0,row{10}={mypink}}
\toprule
\SetCell[r=2]{l} \textbf{Method} &
\SetCell[c=5]{c} \textbf{Single-Subject} & & & & &
\SetCell[c=5]{c} \textbf{Multi-Subject} & & & & &
\SetCell[r=2]{c} \textbf{Overall$\uparrow$} \\
\hline
& DPG$\uparrow$ & ID-S$\uparrow$ & IP-S$\uparrow$ & AES$\uparrow$ & Avg$\uparrow$ &
DPG$\uparrow$ & ID-S$\uparrow$ & IP-S$\uparrow$ & AES$\uparrow$ & Avg$\uparrow$ & \\ 
\midrule
UNO~\cite{UNO}          & 96.04 & 52.16 & 67.10 & \textbf{57.89} & 68.30 % single
             & 88.62 & 35.06 & 59.02 & \underline{54.66} & 59.34 % multi
             & 63.82 \\ % overall
DreamO~\cite{DreamO}       & \textbf{97.19} & 75.48 & 66.91 & 56.27 & 73.95 % single
             & \underline{89.73} & 52.53 & 61.77 & 53.85 & 64.47 % multi
             & 69.21 \\ % overall
OmniGen~\cite{OmniGen}      & 90.61 & 76.63 & 69.86 & 54.70 & 72.95 % single
             & 87.67 & \textbf{74.34} & 57.18 & 53.45 & \underline{68.16} % multi
             & 70.56 \\ % overall
OmniGen2~\cite{OmniGen2}     & \underline{96.65} & 60.76 & 66.53 & 53.31 & 69.31 % single
             & \textbf{91.29} & 38.48 & 60.81 & 52.56 & 60.79 % multi
             & 65.05 \\ % overall
\mbox{MS-Diffusion~\cite{MS-Diffusion}} & 89.20 & 47.42 & \underline{70.28} & 56.67 & 65.89 % single
             & 80.06 & 24.70 & 51.17 & \textbf{54.83} & 52.69 % multi
             & 59.29 \\ % overall
\mbox{MIP-Adapter~\cite{MIP-Adapter}}  & 80.04 & 39.22 & 65.95 & 54.15 & 59.84 % single
             & 83.60 & 21.04 & 49.61 & 53.43 & 51.92 % multi
             & 55.88 \\ % overall
XVerse~\cite{XVerse}       & 92.52 & \underline{79.80} & 67.68 & \underline{57.43} & \underline{74.36} % single
             & 87.40 & \underline{67.15} & \underline{62.59} & 54.59 & 67.93 % multi
             & \underline{71.15} \\ % overall
\textbf{UniVerse (Ours)}    & 91.93 & \textbf{82.77} & \textbf{75.88} & 55.86 & \textbf{78.14} % single
             & 87.95  &  66.69 & \textbf{71.60} & 54.44 & \textbf{70.18}  % multi
             & \textbf{74.16} \\ % overall
\bottomrule
\end{tblr}
% \vspace{-0.2cm}

\end{table*}

\mypara{Evaluation Metrics} 
We follow two evaluation protocols from previous benchmarks: VLM-as-a-judge and feature-based scores. The prior protocol is used in DreamBench++ and leverages GPT-4o~\cite{GPT-4o} to grade (0 to 1) each generated image and its corresponding inputs. It evaluates models in Concept Preservation (CP), Prompt Fidelity (PF), and their multiplication. 
The latter metric is used in XVerseBench and uses pretrained models to compute the similarity between generated images and their inputs. The metrics include Dense Prompt graph (DPG)~\cite{DPG} measuring the prompt alignment, Identity Similarity (ID-S)~\cite{ID-S} in human identity preservation, Perceptual Similarity (IP-S) with DINOv2~\cite{DINOv2} for object appearance consistency, and Attribute Editing Score (AES) with a SigLIP-based predictor~\cite{AES} for evaluating overall aesthetic quality.
In UniVerseBench, we use IP-S and AES metrics to evaluate single- and multi-subject generation quality. 

\mypara{Baselines}
We compare UniVerse against several state-of-the-art personalized image generation models, including UNO~\cite{UNO}, DreamO~\cite{DreamO}, OmniGen~\cite{OmniGen}, OmniGen2~\cite{OmniGen2}, MS-Diffusion~\cite{MS-Diffusion}, and MIP-Adapter~\cite{MIP-Adapter}, and XVerse~\cite{XVerse}.
For all baselines, we used the default configurations in their respective code repositories for evaluation and generation, or the evaluation configurations specified in their respective papers. For consistency in comparison, we set all models to generate a target image size of 768.

\mypara{Model Architecture and Training Details} For our RCE, we use the pretrained CLIP-L/14-224~\cite{CLIP} to extract reference image and prompt features. The perceiver layer follows XVerse implementation. The MLP layer includes two linear layers with an activation in between and a layer norm at the end. For DiT, we use LoRA~\cite{LoRA} with a rank of 128 to adapt to new conditions. The first stage consists of 10 epochs with a learning rate of $1\x10^{-4}$ and a cosine scheduler. We save the best epoch based on IoU on the validation set. For the second stage, we train for 150K iterations in total: the first 100K steps learn the shared offset, and the remaining 50K steps jointly train the block-wise adaptations. We use a learning rate of $ 5 \times 10^{-6}$ with the AdamW~\cite{AdamW} optimizer and a batch size of 16 across 8 NVIDIA A100 GPUs.

\subsection{Qualitative Results}

We present qualitative results of UniVerse in Figures~\ref{fig:qual_single}, \ref{fig:qual_multi}, and \ref{fig:qual_ours}. In single-subject settings (Figure~\ref{fig:qual_single}), prior methods frequently exhibit concept leakage, incomplete attribute preservation, or identity drift, whereas UniVerse consistently extracts the intended concept and maintains subject fidelity—even in challenging cases requiring disentanglement of similar concepts. 

In multi-subject scenarios (Figure~\ref{fig:qual_multi}), baseline models struggle with leakage, incorrect attribute transfer, and compositional failures, particularly when multiple references overlap in concepts. 
UniVerse reliably separates and preserves each concept, capturing fine-grained attributes and composing them coherently. 

% qualitative comparison with baselines

% \begin{figure}[t]
% \vspace{0.5em}
%     \centering
%     \vspace{-0.5em}
%     \includegraphics[width=\linewidth]{author-kit-CVPR2026-v1-latex-/figures/xverse_compare.pdf}
%     \vspace{-.7cm}
%     \caption{\textbf{Qualitative Comparison with XVerse~\cite{XVerse}}.  Despite being entirely segmentation-free, our method produces more faithful multi-concept compositions and preserves subject attributes more reliably than XVerse, which relies on segmentation masks. 
%     }
%     \label{fig:qual_compare_xverse}
%     \vspace{-1em}
% \end{figure}

Figure~\ref{fig:qual_ours} shows three UniVerse-generated images for each group of reference concepts. Across these examples, UniVerse reliably extracts the relevant visual elements from the reference images and recombines them in coherent new scenes. In addition to preserving the core subjects, UniVerse maintains subtle attributes like clothing details, accessories, and overall appearance cues, and can flexibly reimagine these elements within novel scene compositions.

UniVerse demonstrates robust identity-preserving composition in multi-human scenarios Figure~\ref{fig:human}. Furthermore, our method successfully disentangles and composes not only discrete objects but also abstract, non-object attributes such as pose and material Figure~\ref{fig:pose}. While UniVerse maintains high identity fidelity for up to six objects Figure~\ref{fig:multi_obj}, it faces compositional capacity limits as the object count increases (7–9). In such high-density scenes, the model may exhibit object omission or identity crosstalk.

\subsection{Quantitative Results}

% Explanation of quantitative results based on tables
On XVerseBench (Table~\ref{tab:XVerseBench}), UniVerse achieves the highest overall performance for both single-subject and multi-subject generation. 
In single-subject evaluation, UniVerse demonstrates superior identity preservation (ID-S) and appearance similarity (IP-S), achieving an average score of 78.14, outperforming the second-best model, XVerse, by over 3 points. 
In multi-subject scenarios, UniVerse shows strong cross-image compositional abilities, surpassing all baselines by over 2 points. 
While single-subject generation is not the main focus of UniVerse, we achieve competitive performance. We show the result on Dreambench++ in the supplementary material.

% showing results that are largely consistent with other strong baselines on DreamBench++ (Table~\ref{tab:DreamBench++}).
% On DreamBench++ (Table~\ref{tab:DreamBench++}), UniVerse achieves competitive performance, showing results that are largely consistent with other strong baselines in single-subject generation. 

On UniVerseBench (Table~\ref{tab:UniVerseBench}), UniVerse consistently surpasses baseline models in both single- and multi-subject evaluations. 
These results demonstrate UniVerse’s strong generalization across personalization scenarios, achieving state-of-the-art performance in compositional generation while preserving fine-grained visual fidelity and semantic coherence.

\subsection{Ablation Studies}

We conducted a thorough ablation study to validate the contributions of the key components in our UniVerse model, with results summarized in Table \ref{tab:ablation}. 
Our baseline model achieves strong performance on the multi-subject UniVerseBench, with an average score of 48.64. When we remove the Reference Condition Extractor (RCE) pretraining stage, the performance drops by 0.75 points, confirming its positive impact. 
Similarly, removing the cross-reference mechanism during training also degrades performance, resulting in a 0.74 point drop. 
The most significant finding is that removing the visual reference latents during inference causes the largest performance decline, with the average score dropping by 1.50 points. These results conclusively demonstrate that all ablated components are integral and beneficial to the model's overall effectiveness.

% \FloatBarrier

% UniVerseBench
\begin{table}[t]
\caption{Quantitative comparison of single-subject and multi-subject personalization performance on our UniVerseBench. \textbf{Bold} represents best performance, \underline{underline} represents second best.
}
\vspace{-0.3cm}
\label{tab:UniVerseBench}
\footnotesize
\centering
\begin{tblr}{width=\linewidth,colspec={@{}X[2.5,l]|*{3}{X[0.8,c]}|*{3}{X[0.8,c]}|X[0.9,c]@{}},stretch=0,colsep=2pt,row{10}={mypink}}
\toprule
\SetCell[r=2]{l} \textbf{Method} &
\SetCell[c=3]{c} \textbf{Single-Subject} & & &
\SetCell[c=3]{c} \textbf{Multi-Subject} & & &
\SetCell[r=2]{c} Overall$\uparrow$ \\
\hline
& IP-S$\uparrow$ & AES$\uparrow$ & Avg$\uparrow$ & IP-S$\uparrow$ & AES$\uparrow$ & Avg$\uparrow$ & \\ 
\midrule
UNO~\cite{UNO} & 39.92 & \textbf{56.08} & 48.00 % single2
             & 37.91 & \textbf{55.43} & 46.67 % multi
             & 47.19 \\ % overall
DreamO~\cite{DreamO} & 45.49 & 54.16 & 49.83 % single2
             & 39.99 & 54.35 & 47.17 % multi
             & 48.97 \\ % overall
OmniGen~\cite{OmniGen} & 47.72 & 52.64 & 50.18 % single2
             & \underline{41.75} & 54.67 & \underline{48.21} % multi
             & 49.53 \\ % overall
OmniGen2~\cite{OmniGen2} & 44.99 & 48.94 & 46.97 % single2
             & 40.55 & 52.64 & 46.60 % multi
             & 46.50 \\ % overall
\mbox{MS-Diffusion~\cite{MS-Diffusion}} & \underline{50.42} & 53.02 & \underline{51.72} % single2
             & 40.98 & 54.72 & 47.85 % multi
             & \underline{50.49} \\ % overall
\mbox{MIP-Adapter~\cite{MIP-Adapter}} & 45.96 & 48.44 & 47.20 % single2
             & 37.46 & 51.05 & 44.26 % multi
             & 46.31 \\ % overall
XVerse~\cite{XVerse} & 47.11 & \underline{55.89} & 51.50 % single2
             & 40.07 & 51.24 & 45.67 % multi
             & 48.59 \\ % overall
\textbf{UniVerse (Ours)} &  \textbf{51.49} & 54.62 & \textbf{53.06} % single2
             & \textbf{42.29} & \underline{54.98} & \textbf{48.64} % multi
             & \textbf{51.05} \\ % overall
\bottomrule
\end{tblr}
\vspace{-0.7em}
\end{table}

\begin{table}[t]
\centering
\caption{Ablation studies on UniVerse components with our UniVerseBench multiple-object evaluation. We test our model without the Reference Condition Extractor (RCE) pretraining stage and without cross-reference during training. We also measure the effect of omitting visual-reference latents during inference.}
\vspace{-0.3cm}
\label{tab:ablation}
\footnotesize
\begin{tblr}{width=\linewidth,colspec={@{}X[2,l]|*{3}{X[1,c]}X[1,c]},stretch=0,row{2}={mypink},colsep=2pt}
\toprule
\textbf{Settings} & IP-S $\uparrow$ & AES $\uparrow$ & Avg $\uparrow$ & $\Delta$ Avg \\
\midrule
Baseline & 42.29 & 54.98 & 48.64 & 0.00 \\
\SetCell[c=4]{l} \textit{Different training strategy} \\
No RCE Pretraining & 41.82 & 53.96 & 47.89 & \ndif{0.75} \\
No Cross-Reference &41.95 & 53.84 & 47.90 & \ndif{0.74}\\
\SetCell[c=4]{l} \textit{Visual condition in personalization} \\
No Visual Latents & 40.15 & 54.12 & 47.14 & \ndif{1.50}\\
\bottomrule
\end{tblr}
\vspace{-1em}
\end{table}

%% file: sec/5_conclusion.tex
\section{Discussions}
\label{sec:conclusion}
\vspace{-0.5em}

Our method has several limitations. 
%First, we lack a comprehensive, segmentation-free benchmark for multi-reference generation; a future benchmark with richer reference sets (e.g., 3+ concepts with varied attributes) is needed for rigorous evaluation. 
First, a broader challenge in the field is a lack of a comprehensive segmentation-free benchmark for multi-reference generation; a future benchmark with richer reference sets (\eg, 3+ concepts each with multiple attributes) would enable more rigorous evaluation. 
Second, our model is not fully robust to concept interference (leakage), though restrictive prompts like ``just the cat" help to mitigate this. 
Our method also occasionally overfits to a reference subject, and performance degrades when prompts are vague or nonsensical.

In this paper, we presented UniVerse, a unified modulation framework designed to address a critical limitation in personalized visual understanding: the inability to localize and disentangle concepts within multi-object scenes.
Our approach successfully moves beyond the need for segmentation-based supervision, enabling robust, segmentation-free personalization within diffusion transformers. We demonstrated that UniVerse can not only customize generative outputs but also precisely localize target concepts, learning to compose complex scenes and decompose them into their constituent parts. 
Our extensive experiments show that UniVerse significantly outperforms state-of-the-art baselines in both localization accuracy and visual fidelity. 
By enabling decomposable concept extraction even in cluttered images, our work paves the way for more flexible, interpretable, and controllable personalized generation.